\documentclass[runningheads]{llncs}
\usepackage{graphicx}
\usepackage{cite}
\usepackage{xcolor}
\usepackage{bm}
\usepackage{amsmath}
\usepackage{array}
\usepackage{multirow}
\usepackage{algorithm}
\usepackage{algpseudocode}
\usepackage{amssymb}

\usepackage{centernot}

\newcommand{\CI}{\mathrel{\perp\mspace{-10mu}\perp}}

\DeclareMathOperator{\E}{\mathbb{E}}
\long\def\/*#1*/{}

\begin{document}
\title{Semi-supervised Learning for Quantification of Pulmonary Edema in Chest X-Ray Images}
\titlerunning{Semi-supervised Learning for Chest X-Ray Images}
%
\author{Ruizhi Liao\inst{1} \and
Jonathan Rubin\inst{2} \and
Grace Lam\inst{1} \and
Seth Berkowitz\inst{3} \and
Sandeep Dalal\inst{2} \and
William Wells\inst{1,4} \and
Steven Horng\inst{3} \and
Polina Golland\inst{1} } 
\authorrunning{R. Liao et al.}
%
\institute{Massachusetts Institute of Technology, Cambridge, MA, USA \\
\email{ruizhi@mit.edu} \and
Philips Research North America, Cambridge, MA, USA \and
Beth Israel Deaconess Medical Center, Harvard Medical School, Boston, MA, USA \and
Brigham and Women's Hospital, Harvard Medical School, Boston, MA, USA}
\maketitle              
\begin{abstract}
We propose and demonstrate machine learning algorithms to assess the severity of pulmonary edema in chest x-ray images of congestive heart failure patients. Accurate assessment of pulmonary edema in heart failure is critical when making treatment and disposition decisions. Our work is grounded in a large-scale clinical dataset of over 300,000 x-ray images with associated radiology reports. While edema severity labels can be extracted unambiguously from a small fraction of the radiology reports, accurate annotation is challenging in most cases. To take advantage of the unlabeled images, we develop a Bayesian model that includes a variational auto-encoder for learning a latent representation from the entire image set trained jointly with a regressor that employs this representation for predicting pulmonary edema severity. Our experimental results suggest that modeling the distribution of images jointly with the limited labels improves the accuracy of pulmonary edema scoring compared to a strictly supervised approach. To the best of our knowledge, this is the first attempt to employ machine learning algorithms to automatically and quantitatively assess the severity of pulmonary edema in chest x-ray images.

\end{abstract}
\section{Introduction}

We propose and demonstrate a semi-supervised learning algorithm to support clinical decisions in congestive heart failure (CHF) by quantifying pulmonary edema. Limited ground truth labels are one of the most significant challenges in medical image analysis and many other machine learning applications in healthcare. It is of great practical interest to develop machine learning algorithms that take advantage of the entire data set to improve the performance of strictly supervised classification or regression methods. In this work, we develop a Bayesian model that learns probabilistic feature representations from the entire image set with limited labels for predicting edema severity. 

Chest x-ray images are commonly used in CHF patients to assess pulmonary edema, which is one of the most direct symptoms of CHF~\cite{mahdyoon1989radiographic}. CHF causes pulmonary venous pressure to elevate, which in turn causes the fluid to leak from the blood vessels in the lungs into the lung tissue. The excess fluid in the lungs is called pulmonary edema. Heart failure patients have extremely heterogenous responses to treatment~\cite{francis2014heterogeneity}. The assessment of pulmonary edema severity will enable clinicians to make better treatment plans based on prior patient responses and will facilitate clinical research studies that require quantitative phenotyping of the patient status~\cite{chakko1991clinical}. While we focus on CHF, the quantification of pulmonary edema is also useful elsewhere in medicine. Quantifying pulmonary edema in a chest x-ray image could be used as a surrogate for patient intravascular volume status, which would rapidly advance research in sepsis and other disease processes where volume status is critical.

Quantifying pulmonary edema is more challenging than detection of pathologies in chest x-ray images~\cite{wang2017chestx, rajpurkar2017chexnet} because grading of pulmonary edema severity relies on much more subtle image findings (features). Accurate grading of the pulmonary edema severity is challenging for medical experts as well~\cite{hammon2014improving}. Our work is grounded in a large-scale clinical dataset that includes approximately 330,000 frontal view chest x-ray images and associated radiology reports, which serve as the source of the severity labels. Of these, about 30,000 images are of CHF patients. Labels extracted from radiology reports via keyword matching are available for about 6,000 images. Thus our image set includes a large number of images, but only a small fraction of images is annotated with edema severity labels.

We use variational auto-encoder (VAE) to capture the image distribution from both unlabeled and labeled images to improve the accuracy of edema severity grading. Auto-encoder neural networks have shown promise for representational modeling~\cite{kingma2013auto}. Earlier work attempted to learn a separate VAE for each label category from unlabeled and labeled data~\cite{kingma2014semi}. We argue and demonstrate in our experiments that this structure does not fit our application well, because pulmonary edema severity score is based on subtle image features and should be represented as a continuous quantity. Instead, we learn one VAE from the entire image set. By training the VAE jointly with a regressor, we ensure it captures compact feature representations for scoring pulmonary edema severity. Similar setups have also been employed in computer vision~\cite{kamnitsas2018semi}. The experimental results show that our method outperforms the multi-VAE approach~\cite{kingma2014semi}, the entropy minimization based self-learning approach~\cite{grandvalet2005semi}, and strictly supervised learning. To the best of our knowledge, this paper demonstrates the first attempt to employ machine learning algorithms to automatically and quantitatively assess the severity of pulmonary edema from chest x-ray images. 

\section{Methods}
\label{sec:methods}

\begin{figure*}[t]
\centerline{
\includegraphics[width=0.8\textwidth]{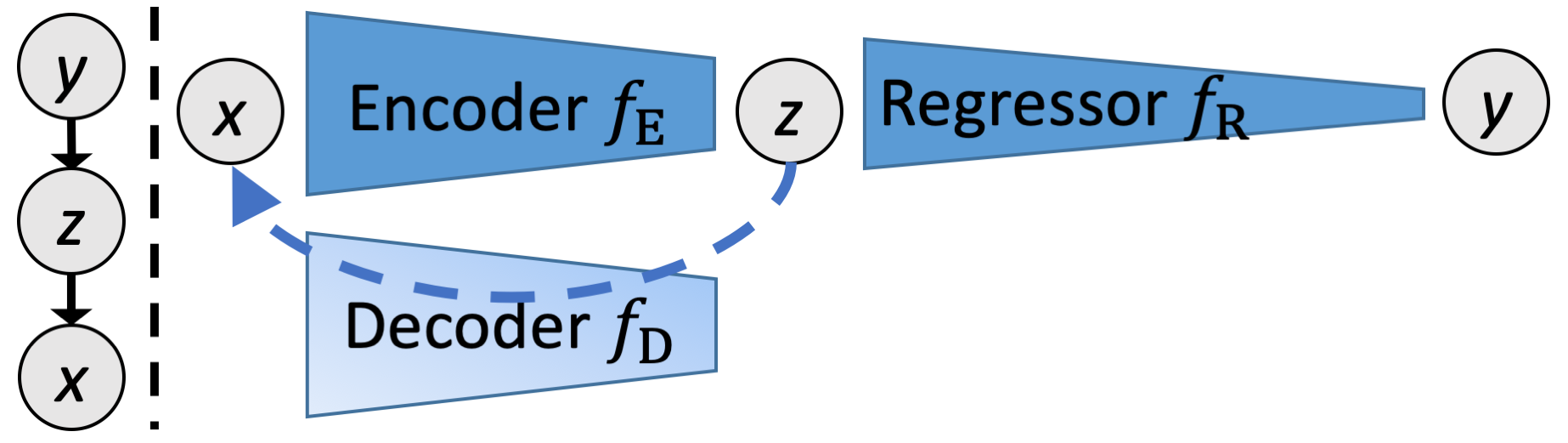}
}
\centerline{
\hskip0.4in
(a)
\hskip1.65in
(b)
\hskip 2.0in
}
\caption{Graphical model and inference algorithm. (a): Probabilistic graphical model, where $x$ represents chest x-ray image, $z$ represents latent feature representation, and $y$ represents pulmonary edema severity. (b): Our computational model. We use neural networks for implementing the encoder, decoder, and regressor. The dashed line (decoder) is used in training only. The network architecture is provided in the supplementary material.}
\label{fig:model_illustration}
\end{figure*}

Let $x\in\mathbb{R}^{n\times{n}}$ be a 2D x-ray image and $y\in{\{0,1,2,3\}}$ be the corresponding edema severity label. Our dataset includes a set of $N$ images $\mathbf{x}=\{x_{i}\}_{i=1}^{N}$ with the first $N_{\text{L}}$ images annotated with severity labels $\mathbf{y}=\{y_{i}\}_{i=1}^{N_\text{L}}$. Here, we derive a learning algorithm that constructs a compact probabilistic feature representation $z$ that is learned from all images and used to predict pulmonary edema severity. Fig.~\ref{fig:model_illustration} illustrates the Bayesian model and the inference algorithm.

\paragraph{\textbf{Learning.}}
The learning algorithm maximizes the log probability of the data with respect to parameters~$\theta$:
\begin{align}
\log p(\mathbf{x}, \mathbf{y}; \theta)=\sum_{i=1}^{N_\text{L}} \log p(\text{x}_{i}, \text{y}_{i}; \theta) + \sum_{i=N_\text{L}+1}^{N} \log p(\text{x}_{i}; \theta).
\label{eq:joint_factorization}
\end{align}
We model $z$ as a continuous latent variable with a prior distribution~$p(z)$, which generates images and predicts pulmonary edema severity. Unlike~\cite{kingma2014semi} that constructs a separate encoder $q(z|x,y)$ for each value of discrete label $y$, we use a single encoder $q(z|x)$ to capture image structure relevant to labels. Distribution $q(z|x)$ serves as a variational approximation for $p(z|x, y)$ for the lower bound:
\begin{align}
\mathcal{L}_{1}({\theta}; \text{x}_{i}, \text{y}_{i}) = & \log p(\text{x}_{i}, \text{y}_{i}; \theta) - D_{KL}(q(\text{z}_{i}|\text{x}_{i}; \theta)||p(\text{z}_{i}|\text{x}_{i}, \text{y}_{i})), \nonumber \\
= & \E_{q(\text{z}_{i}|\text{x}_{i}; \theta)}\big[\log p(\text{x}_{i}, \text{y}_{i}; \theta) + \log p(\text{z}_{i}|\text{x}_{i}, \text{y}_{i}) - \log q(\text{z}_{i}|\text{x}_{i}; \theta) \big] \nonumber \\
= & \E_{q(\text{z}_{i}|\text{x}_{i}; \theta)}\big[\log p(\text{x}_{i}, \text{y}_{i}| \text{z}_{i}; \theta) + \log p(\text{z}_{i}) - \log q(\text{z}_{i}|\text{x}_{i}; \theta) \big]\nonumber \\
=  & \E_{q(\text{z}_{i}|\text{x}_{i}; \theta)}\big[\log p(\text{x}_{i}, \text{y}_{i}| \text{z}_{i}; \theta)\big] - D_{KL}(q(\text{z}_{i}|\text{x}_{i}; \theta)||p(\text{z}_{i})). \nonumber
\end{align}
We assume that $x$, $z$, and $y$ form a Markov chain, i.e., $y \CI x \mid z$, and therefore
\begin{align}
\mathcal{L}_{1}({\theta}; \text{x}_{i}, \text{y}_{i})
=  &  \E_{q(\text{z}_{i}|\text{x}_{i}; \theta_{\text{E}})}\big[\log p(\text{x}_{i}|\text{z}_{i}; \theta_{\text{D}})\big] + \E_{q(\text{z}_{i}|\text{x}_{i}; \theta_{\text{E}})}\big[\log p(\text{y}_{i}|\text{z}_{i}; \theta_{\text{R}}) \big]\nonumber \\ 
& -D_{KL}(q(\text{z}_{i}|\text{x}_{i}; \theta_{\text{E}})||p(\text{z}_{i})),
\label{eq:bound_labelled}
\end{align}
where $\theta_{\text{E}}$ are the parameters of the encoder, $\theta_{\text{D}}$ are the parameters of the decoder, and $\theta_{\text{R}}$ are the parameters of the regressor. Similarly, we have a variational lower bound for~$\log p(\text{x}_{i}; \theta)$:
\begin{align}
\mathcal{L}_{2}({\theta}; \text{x}_{i})
=  \E_{q(\text{z}_{i}|\text{x}_{i}; \theta_{\text{E}})}\big[\log p(\text{x}_{i}|\text{z}_{i}; \theta_{\text{D}})\big]-D_{KL}(q(\text{z}_{i}|\text{x}_{i}; \theta_{\text{E}})||p(\text{z}_{i})).
\label{eq:bound_unlabelled}
\end{align}

By substituting Eq.~(\ref{eq:bound_labelled}) and Eq.~(\ref{eq:bound_unlabelled}) into Eq.~(\ref{eq:joint_factorization}), we obtain a lower bound for the log probability of the data and aim to minimize the negative lower bound:
\begin{align}
\mathcal{J}({\theta}; \mathbf{x}, \mathbf{y})
= & - \sum_{i=1}^{N_\text{L}} \mathcal{L}_{1}({\theta}; \text{x}_{i}, \text{y}_{i}) - \sum_{i=N_\text{L}+1}^{N} \mathcal{L}_{2}({\theta}; \text{x}_{i}) \nonumber \\
= & \sum_{i=1}^{N}{D_{KL}(q(\text{z}_{i}|\text{x}_{i}; \theta_{\text{E}})||p(\text{z}_{i}))}
-\sum_{i=1}^{N_\text{L}}{ \E_{q(\text{z}_{i}|\text{x}_{i}; \theta_{\text{E}})}\big[\log p(\text{y}_{i}|\text{z}_{i}; \theta_{\text{R}})\big]} \nonumber \\
& -\sum_{i=1}^{N}{ \E_{q(\text{z}_{i}|\text{x}_{i}; \theta_{\text{E}})}\big[\log p(\text{x}_{i}|\text{z}_{i}; \theta_{\text{D}})\big]}.
\label{eq:objective_function}
\end{align}

\paragraph{\textbf{Latent Variable Prior~$p(z)$.}}
We let the latent variable prior~$p(z)$ be a multivariate normal distribution, which serves to regularize the latent representation of images. 

\paragraph{\textbf{Latent Representation~$q(z|x)$.}}
We apply the reparameterization trick used in~\cite{kingma2013auto}. Conditioned on image~$\text{x}_i$, the latent representation becomes a multivariate Gaussian variable, $\text{z}_{i}|\text{x}_{i}\sim \mathcal{N}(\text{z}_{i}; \mu_{i},\,\Lambda_{i})$, where~$\mu_{i}$ is a $D$-dimensional vector~$[\mu_{ik}]_{k=1}^{D}$ and~$\Lambda_{i}$ is a diagonal covariance matrix represented by its diagonal elements as~$[\lambda_{ik}^2]_{k=1}^{D}$. Thus, the first term in Eq.~(\ref{eq:objective_function}) becomes:
\begin{align}
\mathcal{J}_{KL}(\theta_{\text{E}}; \text{x}_{i}) = -\frac{1}{2} \sum_{k=1}^{D} \left(\log \lambda_{ik}^2 - \mu_{ik}^2 - \lambda_{ik}^2 \right) + \text{const.}
\label{eq:ae_kl}
\end{align}
We implement the encoder as a neural network~$f_\text{E}(x; \theta_{\text{E}})$ that estimates the mean and the variance of~$z|x$. Samples of~$z$ can be readily generated from this estimated Gaussian distribution. We use one sample per image for training the model.

\paragraph{\textbf{Ordinal Regression~$p(y|z)$.}}
In radiology reports, pulmonary edema severity is categorized into four groups: no/mild/ moderate/severe. Our goal is to assess the severity of pulmonary edema as a continuous quantity. We employ ordinal representation to capture the ordering of the categorical labels. We use a 3-bit representation $\text{y}_{i}=[\text{y}_{ij}]_{j=1}^{3}$ for the four severity levels. The three bits represent the probability of any edema, of moderate or severe edema, and of severe edema respectively (i.e., ``no" is~$[0,0,0]$, ``mild" is~$[1,0,0]$, ``moderate" is~$[1,1,0]$, and ``severe" is~$[1,1,1]$). This encoding yields probabilistic output, i.e., both the estimate of the edema severity and also uncertainty in the estimate. The three bits are assumed to be conditionally independent given the image:
\begin{equation}
p(\text{y}_{i}|\text{z}_{i}; \theta_{\text{R}}) = \prod_{j=1}^{3} f_\text{R}^{j}(\text{z}_{i}; \theta_{\text{R}})^{\text{y}_{ij}}\left(1-f_\text{R}^{j}(\text{z}_{i}; \theta_{\text{R}})\right)^{1-{\text{y}_{ij}}}, \nonumber \\
\end{equation}
where $\text{y}_{ij}$ is a binary label and $f_{\text{R}}^{j}(\text{z}_{i}; \theta_{\text{R}})$ is interpreted as the conditional probability $p(\text{y}_{ij}=1|\text{z}_{i})$. $f_\text{R}(\cdot)$ is implemented as a neural network. The second term in Eq.~(\ref{eq:objective_function}) becomes the cross entropy:
\begin{align}
\mathcal{J_\text{R}}(\theta_{\text{E}}, \theta_{\text{R}}; \text{y}_{i}, \text{z}_{i}) 
= & - \sum_{j=1}^{3} {\text{y}_{ij}}\log f_\text{R}^{j}(\text{z}_{i}; \theta_{\text{R}}) 
- \sum_{j=1}^{3} (1-\text{y}_{ij})\log \left(1-f_\text{R}^{j}(\text{z}_{i}; \theta_{\text{R}})\right).
\label{eq:clf_distribution}
\end{align}

\paragraph{\textbf{Decoding~$p(x|z)$.}}
We assume that image pixels are conditionally independent (Gaussian) given the latent representation. Thus, the third term in Eq.~(\ref{eq:objective_function}) becomes: 
\begin{align}
\mathcal{J_\text{D}}(\theta_{\text{E}}, \theta_{\text{D}}; \text{x}_{i}, \text{z}_{i})
& = - \log \mathcal{N}(\text{x}_{i}; f_{\text{D}}(\text{z}_{i}; \theta_{\text{D}}), \Sigma_{i}) \nonumber \\
& = \dfrac{1}{2}(\text{x}_{i}-f_{\text{D}}(\text{z}_{i}; \theta_{\text{D}}))^{T}\Sigma_{i}^{-1}(\text{x}_{i}-f_{\text{D}}(\text{z}_{i}; \theta_{\text{D}}))+\text{const.},
\label{eq:ae_distribution}
\end{align}
where $f_\text{D}(\cdot)$ is a neural network decoder that generates an image implied by the latent representation~$z$, and $\Sigma_{i}$ is a diagonal covariance matrix. 

\paragraph{\textbf{Loss Function.}}
Combining Eq.~(\ref{eq:ae_kl}), Eq.~(\ref{eq:clf_distribution}) and Eq.~(\ref{eq:ae_distribution}), we obtain the loss function for training our model:
\begin{align}
\mathcal{J} (\theta_{\text{E}}, \theta_{\text{R}}, \theta_{\text{D}}; \mathbf{x}, \mathbf{y})
= & \sum_{i=1}^{N} \mathcal{J}_{KL}(\theta_{\text{E}}; \text{x}_{i}) + \sum_{i=1}^{N_\text{L}} \mathcal{J_\text{R}}(\theta_{\text{E}}, \theta_{\text{R}}; \text{y}_{i}, \text{z}_{i}) + \sum_{i=1}^{N} \mathcal{J_\text{D}}(\theta_{\text{E}}, \theta_{\text{D}}; \text{x}_{i}, \text{z}_{i}) \nonumber \\
= & -\frac{1}{2} \sum_{i=1}^{N} \sum_{k=1}^{D} \left(\log \lambda_{ik}^2 - \mu_{ik}^2 - \lambda_{ik}^2 \right) \nonumber \\
& - \sum_{i=1}^{N_\text{L}} \left( \sum_{j=1}^{3} {\text{y}_{ij}}\log f_\text{R}^{j}(\text{z}_{i}; \theta_{\text{R}}) + \sum_{j=1}^{3} (1-\text{y}_{ij})\log \left(1-f_\text{R}^{j}(\text{z}_{i}; \theta_{\text{R}})\right) \right) \nonumber \\
& + \dfrac{1}{2} \sum_{i=1}^{N} \left(\text{x}_{i}-f_{\text{D}}(\text{z}_{i}; \theta_{\text{D}}))^{T}\Sigma_{i}^{-1}(\text{x}_{i}-f_{\text{D}}(\text{z}_{i}; \theta_{\text{D}})\right).
\label{eq:loss_function}
\end{align}
We employ the stochastic gradient-based optimization procedure Adam~\cite{kingma2014adam} to minimize the loss function. Our training procedure is outlined in the supplementary materiel. The pulmonary edema severity category extracted from radiology reports is a discrete approximation of the actual continuous severity level. To capture this, we compute the expected severity:
\begin{align}
\hat{y}
 =0 \times (1-\hat{y}_1) + 1\times (\hat{y}_1-\hat{y}_2)+2\times (\hat{y}_2-\hat{y}_3) +3 \times \hat{y}_3 =\hat{y}_1+\hat{y}_2+\hat{y}_3. \nonumber
\end{align}

\section{Implementation Details}
The size of the chest x-ray images in our dataset varies and is around 3000$\times$3000 pixels. We randomly rotate and translate the images (differently at each epoch) on the fly during training and crop them to 2048$\times$2048 pixels as part of data augmentation. We maintain the original image resolution to preserve subtle differences between different levels of pulmonary edema severity.

The encoder is implemented as a series of residual blocks~\cite{he2016deep}. The decoder is implemented as a series of transposed convolutional layers, to build an output image of the same size as the input image (2048$\times$2048). The regressor is implemented as a series of residual blocks with an averaging pooling layer followed by two fully connected layers. The regressor output $\hat{y}$ has 3 channels. The latent representation~$z$ has a size of 128$\times$128. During training, one sample is drawn from~$z$ per image. The KL-loss (Eq.~(\ref{eq:ae_kl})) and the image reconstruction error (Eq.~(\ref{eq:ae_distribution})) in the loss function are divided by the latent feature size and the image size respectively. The variances in Eq.~(\ref{eq:ae_distribution}) are set to 10, which gives a weight of 0.1 to the image reconstruction error. The learning rate for the Adam optimizer training is 0.001 and the minibatch size is 4. The model is trained on a training dataset and evaluated on a separate validation dataset every few epochs during training. The model checkpoint with the lowest error on the validation dataset is used for testing. The neural network architecture is provided in the supplementary material.

\section{Experiments}

\paragraph{\textbf{Data.}}
Approximately 330,000 frontal view x-ray images and their associated radiology reports were collected as part of routine clinical care in the emergency department of Beth Israel Deaconess Medical Center and subsequent in-hospital stay. A subset of the image set has been released~\cite{johnson2019mimic}.

\begin{figure*}[t]
\centerline{
\hfill
\includegraphics[width=1.1 in]{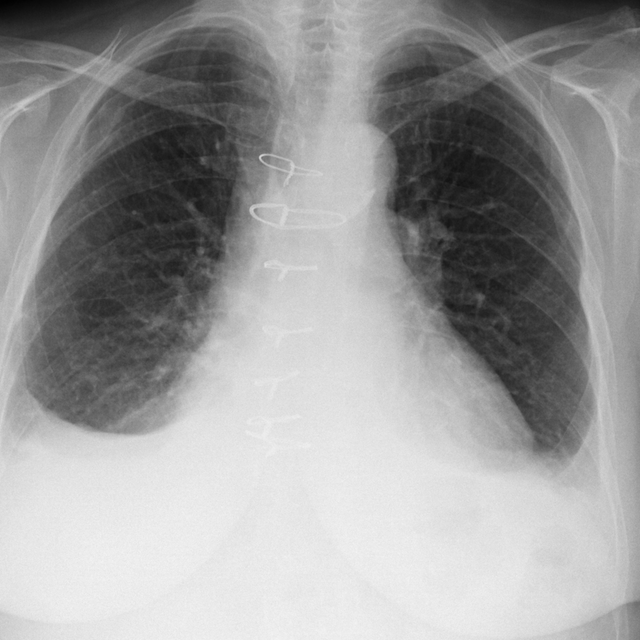}
\hskip0.02in
\includegraphics[width=1.1 in]{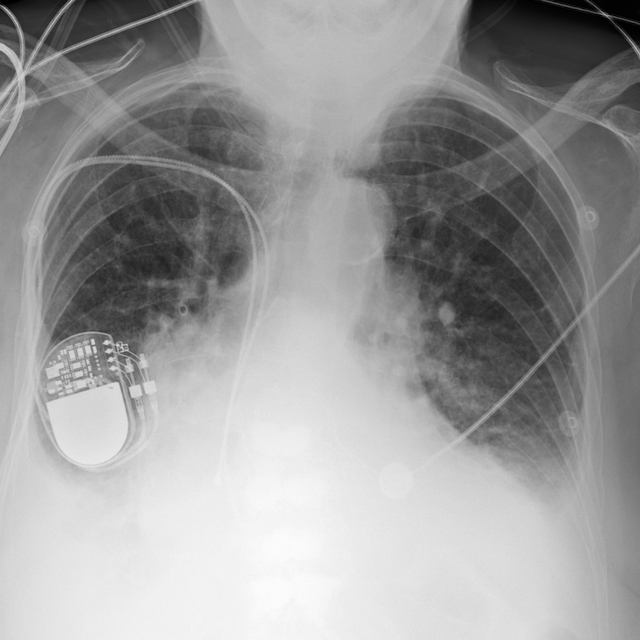}
\hskip0.02in
\includegraphics[width=1.1 in]{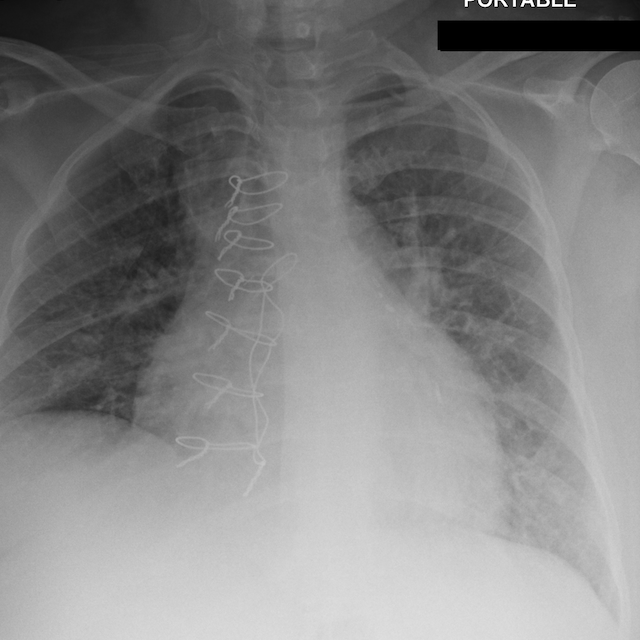}
\hskip0.02in
\includegraphics[width=1.1 in]{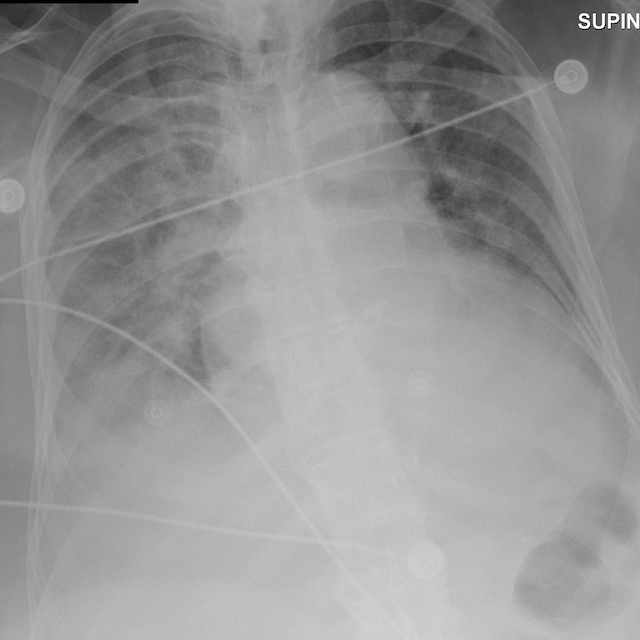}
\hfill
}
\vskip0.03in
\centerline{
\hfill
\begin{minipage}[t]{1.1in}
\centering
No edema
\end{minipage} 
\hskip0.02in
\begin{minipage}[t]{1.1in}
\centering
Mild edema
\end{minipage} 
\hskip0.02in
\begin{minipage}[t]{1.1in}
\centering
Moderate edema
\end{minipage}
\hskip0.02in
\begin{minipage}[t]{1.1in}
\centering
Severe edema
\end{minipage} 
\hfill
}
\caption{Representative chest x-ray images with varying severity of pulmonary edema.}
\vskip-0.1in
\label{fig:example_images}
\end{figure*}

In this work, we extracted the pulmonary edema severity labels from the reports by searching for keywords that are highly correlated with a specific stage of pulmonary edema. Due to the high variability of wording in radiology reports, the same keywords can mean different clinical findings in varying disease context. For example, perihilar infiltrate means moderate pulmonary edema for a heart failure patient, but means pneumonia in a patient with a fever. To extract meaningful labels from the reports using the keywords, we limited our label extraction to a CHF cohort. This cohort selection yielded close to 30,000 images, of which 5,771 images could be labeled via our keyword matching. Representative images of each severity level are shown in Fig.~\ref{fig:example_images}. The data details are summarized in the supplementary material. 

\paragraph{\textbf{Evaluation.}}
We randomly split the images into training (4,537 labeled images, 334,664 unlabeled images), validation (628 labeled images), and test (606 labeled images) image sets. There is no patient overlap between the sets. Unlabeled images of the patients in the validation and test sets are excluded from training. The labeled data split is 80$\%$/10$\%$/10$\%$ into training/validation/test respectively. 

We evaluated four methods: (i)~\textit{Supervised}: purely supervised training that uses labeled images only; (ii)~\textit{EM}: supervised training with labeled images that imputes labels for unlabeled images and minimizes the entropy of predictions~\cite{grandvalet2005semi}; (iii)~\textit{DGM}: Semi-supervised training with deep generative models (multiple VAEs) as described in~\cite{kingma2014semi}; (iv)~\textit{VAE}\_\textit{R}: Our method that learns probabilistic feature representations from the entire image set with limited labels. For the baseline supervised learning method, we investigated different neural network architectures previously demonstrated for chest x-ray images~\cite{wang2017chestx, rajpurkar2017chexnet} and did not find the network architecture changes the supervised learning results significantly.

We evaluate the methods on the test image set using the root mean squared (RMS) error and Pearson correlation coefficient (CC). 

\begin{figure*}[t]
\centerline{
\includegraphics[width=1.3in]{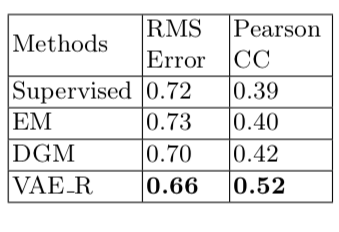}
\hskip0.5in
\includegraphics[width=2.1in]{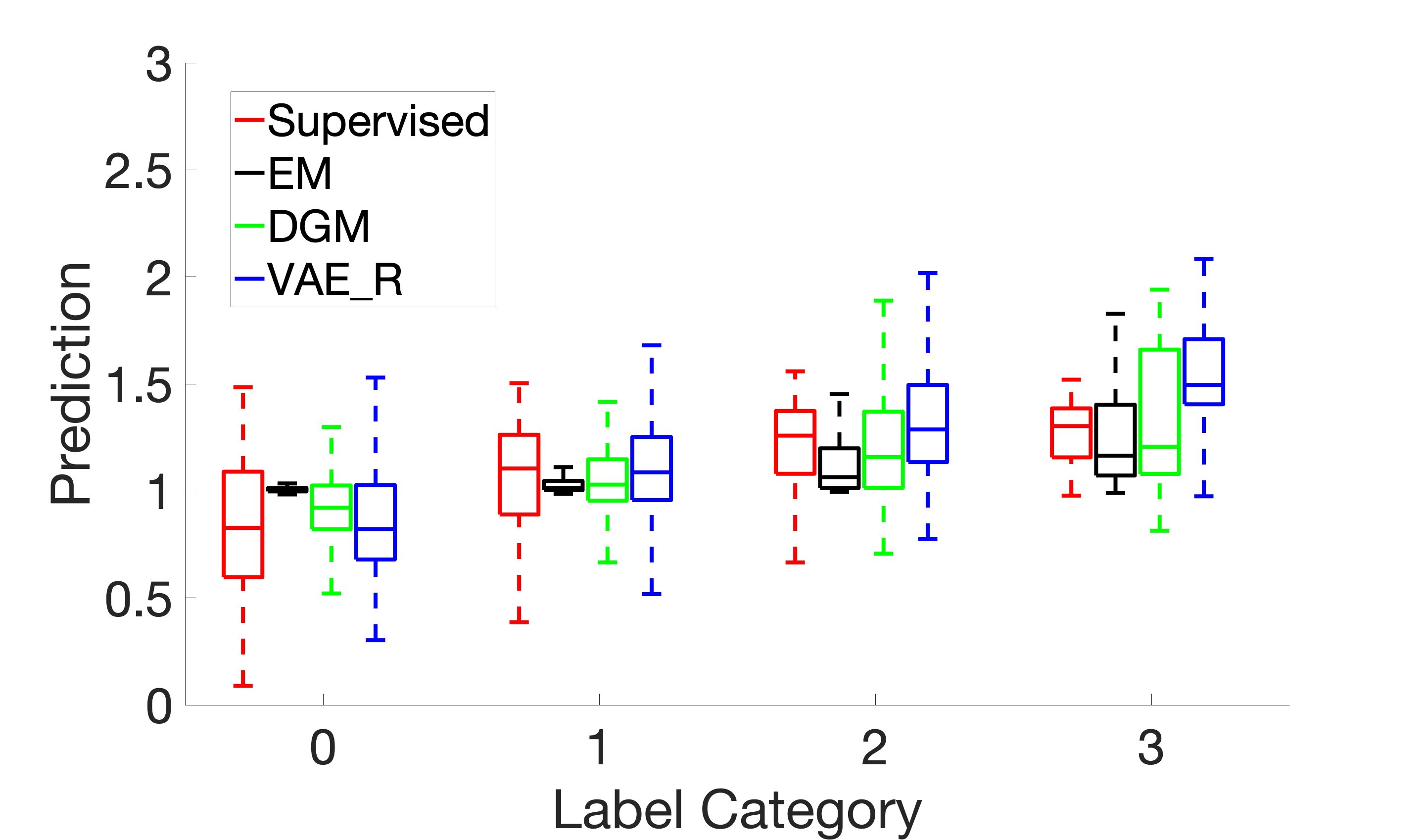}
}
\vskip-0.1in
\caption{Summary of the results. Left table: Root mean squared errors and Pearson correlation coefficients for each method. Right plot: Predicted edema severity scores on each label category.}
\vskip-0.1in
\label{fig:results}
\end{figure*}

\paragraph{\textbf{Results.}}
Fig.~\ref{fig:results} summarizes the prediction performance of the four methods. The method that jointly learns probabilistic feature representations outperforms the other three models.

\section{Conclusions}

In this paper, we demonstrated a regression model augmented with a VAE trained on a large image dataset with a limited number of labeled images. Our results suggest that it is difficult for a generative model to learn distinct data clusters for the labels that rely on subtle image features. In contrast, learning compact feature representations jointly from images and limited labels can help inform prediction by capturing structure shared by the image distribution and the conditional distribution of labels given images. 

We demonstrated the first attempt to employ machine learning algorithms to automatically and quantitatively assess the severity of pulmonary edema from chest x-ray images. Our results suggest that granular information about a patient's status captured in medical images can be extracted by machine learning algorithms, which promises to enable clinicians to deliver better care by quantitatively summarizing an individual patient's medical history, for example response to different treatments. This work also promises to enable clinical research studies that require quantitative summarization of patient status.  

%
%
%
\bibliographystyle{splncs04}

\pagebreak

\hfill \break

\section*{Supplementary Material}

\hfill \break
\hfill \break

\begin{algorithm}
\caption{Stochastic learning with minibatch for the model in Fig.~\ref{fig:model_illustration}}
\label{training_algorithm}
\begin{algorithmic}[100]
\State $\theta_{\text{E}}, \theta_{\text{R}}, \theta_{\text{D}} \gets$ Initialize parameters
\Repeat
\Repeat \Comment{Training on labeled data}
\State $\mathbf{y}^{n}, \mathbf{x}^{n} \gets$ Random minibatch of n image and label pairs
\State $\mathbf{z}^{n} \gets$ Samples from $q(z|\mathbf{x}^{n}; \theta_E)$ (one sample per image)
\State $\text{g} \gets \nabla_{\theta_{\text{E}}}\mathcal{J}_{KL}(\theta_{\text{E}}; \mathbf{x}^{n}) + \nabla_{\theta_{\text{E}}, \theta_{\text{R}}}{\mathcal{J_\text{R}}(\theta_{\text{E}}, \theta_{\text{R}}; \mathbf{y}^{n}, \mathbf{z}^{n})} + \nabla_{\theta_{\text{E}}, \theta_{\text{D}}}{\mathcal{J_\text{D}}(\theta_{\text{E}}, \theta_{\text{D}}; \mathbf{x}^{n}, \mathbf{z}^{n})}$
\State $\theta_{\text{E}}, \theta_{\text{R}}, \theta_{\text{D}} \gets$ Update parameters using gradients $\text{g}$ (e.g., Adam~\cite{kingma2014adam})
\Until{the last minibatch of labeled set}
\Repeat \Comment{Training on unlabeled data}
\State $\mathbf{x}^{n} \gets$ Random minibatch of n images
\State $\mathbf{z}^{n} \gets$ Samples from $q(z|\mathbf{x}^{n}; \theta_E)$ (one sample per image)
\State $\text{g} \gets \nabla_{\theta_{\text{E}}}\mathcal{J}_{KL}(\theta_{\text{E}}; \mathbf{x}^{n}) + \nabla_{\theta_{\text{E}}, \theta_{\text{D}}}{\mathcal{J_\text{D}}(\theta_{\text{E}}, \theta_{\text{D}}; \mathbf{x}^{n}, \mathbf{z}^{n})}$
\State $\theta_{\text{E}}, \theta_{\text{D}} \gets$ Update parameters using gradients $\text{g}$ (e.g., Adam~\cite{kingma2014adam})
\Until{the last minibatch of unlabeled set}
\Until{convergence}
\end{algorithmic}
\end{algorithm}

\hfill \break
\hfill \break
\hfill \break

\begin{table}
\begin{center}
\caption{Summary of chest x-ray image data and pulmonary edema severity labels.}
\label{table:data_labels}
\begin{tabular}{ | m{2.2cm} | m{3cm}| m{1.2cm} | m{2.2cm} | m{1.2cm} | m{2cm} | } 
\hline
Patient cohort &  Edema severity & \multicolumn{2}{l|}{Number of patients} & \multicolumn{2}{l|}{Number of images} \\ 
\hline
\multirow{5}{2.2cm}{Congestive heart failure (CHF)} & No edema & \multicolumn{1}{l}{441} & (0.69\%) & \multicolumn{1}{l}{1003} & (0.29\%)\\ 
& Mild edema & \multicolumn{1}{l}{979} & (1.52\%) & \multicolumn{1}{l}{3058} & (0.89\%)\\
& Moderate edema & \multicolumn{1}{l}{478} & (0.74\%) & \multicolumn{1}{l}{1414} & (0.41\%)\\
& Severe Edema & \multicolumn{1}{l}{116} & (0.18\%) & \multicolumn{1}{l}{296} & (0.09\%)\\
& Unlabeled & \multicolumn{1}{l}{1880} & (2.92\%) & \multicolumn{1}{l}{22021} & (6.40\%)\\
\hline
Others & Unlabeled & \multicolumn{1}{l}{60406} & (93.95\%) & \multicolumn{1}{l}{316479} & (91.92\%)\\
\hline
\end{tabular}
\end{center}
\end{table}

\begin{figure*}
\centerline{
\includegraphics[width=1\textwidth]{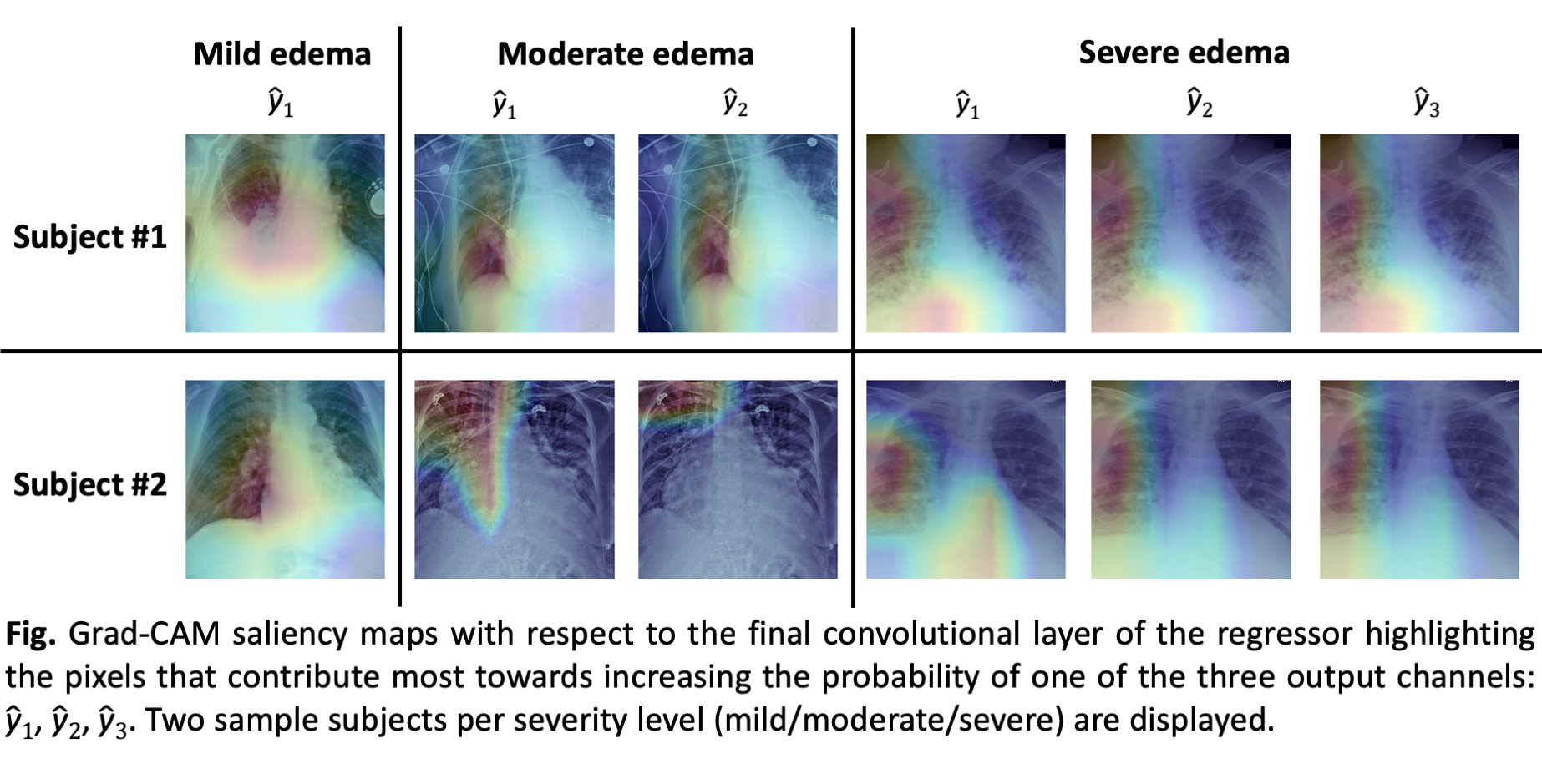}
}
\end{figure*}

\begin{figure*}
\centerline{
\includegraphics[width=1\textwidth]{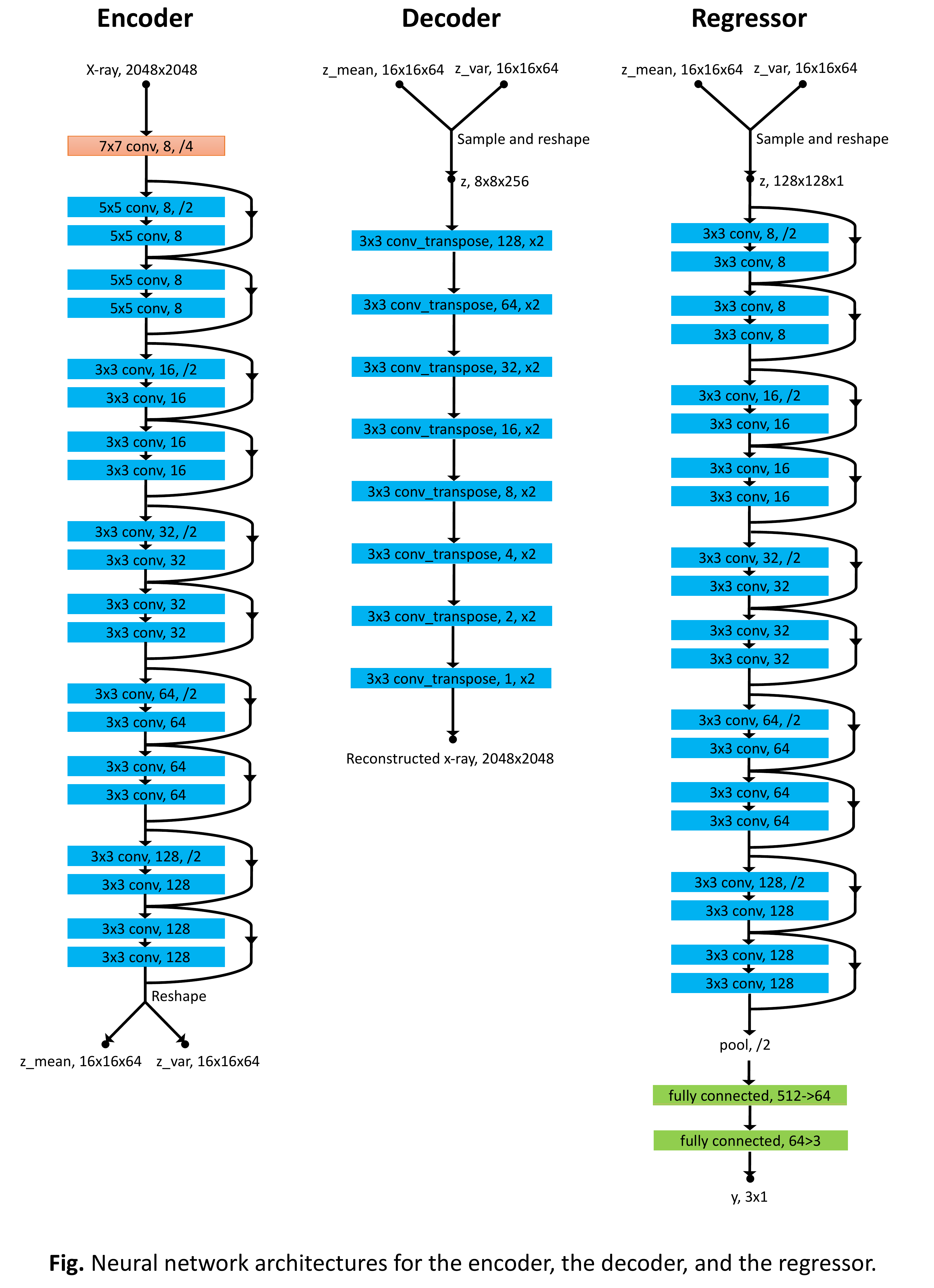}
}
\end{figure*}

\end{document}